\documentclass[11pt]{article}
\usepackage[final]{acl}
\usepackage{times}
\usepackage{latexsym}
\usepackage[T1]{fontenc}
\usepackage[utf8]{inputenc}
\usepackage{microtype}
\IfFileExists{inconsolata.sty}{\usepackage{inconsolata}}{}
\usepackage{booktabs}
\usepackage{amsmath}
\usepackage{xcolor}
\usepackage{tikz}
\usepackage{float}
\newcommand{\pass}{\textsc{Pass}}
\newcommand{\fail}{\textsc{Fail}}
\newcommand{\gpt}{GPT-5.5}
\newcommand{\opus}{Opus 4.7}

\usepackage{placeins}
\hypersetup{
  pdftitle={When Hiring Becomes Agent-Mediated: Evaluating Access and Recurrence in Two-Agent Resume Screening},
  pdfauthor={Jian Gao, Hang Jiang}
}

\title{When Hiring Becomes Agent-Mediated:\\
Evaluating Access and Recurrence in Two-Agent R\'esum\'e Screening}
\author{
  Jian Gao \\
  Northeastern University \\
  Boston, MA, USA
  \And
  Hang Jiang \\
  Northeastern University \\
  Boston, MA, USA
}

\begin{document}
\maketitle

\begin{abstract}
Hiring is bilateral: employers assess fit, while candidates present and defend
evidence of their qualifications. Yet r\'esum\'e screening---the first
gate---is commonly automated as a static, one-call judgment over a
r\'esum\'e--job pair. We study a two-agent alternative in which employer-side
and candidate-side agents represent these roles, exchange evidence, and update
their judgments before deciding who advances. We compare procedures on
600 constructed r\'esum\'e--job pairs using \gpt{} and Claude Opus 4.7. Two-agent
screening advances more applications (33.3\% to 39.3\% for \gpt{};
34.0\% to 35.5\% for \opus{}). Across three runs on the common 191-pair borderline pool,
pass-instance rates rise from 4.5\% to 26.2\% and from 6.5\% to 16.1\%,
respectively. This is not a uniform relaxation: two-agent screening 
rejects applications one-call advances, changing decisions in both directions. At similar pass volumes, the procedures advance different
applications, and no one-call threshold recovers applications consistently
selected by two-agent screening. Among discovery-selected cases
re-executed in fresh runs, two-agent-only selections recur less often than
shared selections---clearly under \gpt{}, less certainly under \opus{}---while a
separate one-call follow-up shows no comparable decline.
As hiring becomes agent-mediated on both sides, the screening
procedure---not only the model behind it---shapes who reaches human review and
how reliably that access recurs.
\end{abstract}

\section{Introduction}

R\'esum\'e screening has traditionally been an employer-controlled gate:
employers review application materials and decide which candidates receive further consideration. Current LLM-based screening largely preserves this
structure, replacing the initial review with a single model call over a fixed
r\'esum\'e--job pair
\citep{an-etal-2024-large,iso-etal-2025-evaluating,
vaishampayan-etal-2025-human}. As agent systems mature, however, employers and
candidates may both delegate parts of this process. A candidate-side agent can
surface and explain evidence already contained in the r\'esum\'e, while an
employer-side agent can evaluate that evidence against the job description and
raise unresolved requirements
\citep{lo-etal-2025-ai,yuksel-etal-2026-agentic}.

This two-agent design is motivated by a division of functions, not by an
assumption that dialogue is inherently better. The candidate-side agent
responds to employer-side concerns using r\'esum\'e-grounded evidence, while
the employer-side agent updates its assessment as those concerns are addressed.
No new applicant information is introduced. What changes is how evaluation and
advocacy are assigned, how evidence is exchanged, and when judgments are
revised before the screening decision. Two-agent screening is therefore a
different decision procedure rather than merely a longer prompt.

A \pass{} does not imply hiring, but it controls access to human review. Pass
rates alone are therefore insufficient: two procedures may advance similar
numbers but different applications, and a procedure-specific selection may
disappear when the pipeline is executed again. The relevant question is not
only whether two-agent screening changes the number of applications advanced,
but how it reallocates access relative to one-call screening and whether those
changes recur.

We compare fixed one-call and two-agent procedures on 600 constructed
r\'esum\'e--job pairs under \gpt{} and \opus{}, treating re-execution as a
first-class measurement. First, two-agent screening changes decisions in both
directions rather than uniformly relaxing one-call screening. Second, even at
similar pass volumes, the procedures advance different applications, and no
threshold on the stored one-call mean reconstructs the persistent two-agent
set. Third, within discovery-selected cohorts, fresh runs show that
two-agent-only selections recur less often than selections shared across
procedures, while a separate one-call follow-up shows no comparable decline.
We compare complete procedures: we neither rank their hiring quality nor
attribute the observed differences to role separation, dialogue, compute, the
decision rules, or any single component.

\section{Related Work}

LLM-based hiring research has examined discrimination, matching quality, and
human--LLM agreement, largely by treating screening as a one-call assessment of
a jointly presented r\'esum\'e and job description
\citep{an-etal-2024-large,iso-etal-2025-evaluating,
vaishampayan-etal-2025-human}. Recent systems demonstrate agentic candidate
assessment and multi-agent r\'esum\'e screening
\citep{lo-etal-2025-ai,yuksel-etal-2026-agentic}. We study the transition
between these settings: when employer evaluation and candidate advocacy are
assigned to separate agents, does the resulting procedure preserve which
applications advance?

Our evaluation connects three reliability traditions. Predictive multiplicity
and prediction churn show that similar aggregate behavior can mask different
instance-level outcomes
\citep{marx2020predictive,black2022model,milanifard2016launch}. Work on LLM and
agent reliability emphasizes repeated execution, dependable long-horizon
behavior, and reproducible reporting
\citep{atil2025nondeterminism,song-etal-2025-good,gupta2026reliabilitybench,
ma2026maestro,rabanser2026science,pattnayak-bhatia-2026-reproevalcard}.
Relatedly, \citet{jiang-etal-2024-personallm} study whether
persona-conditioned agents consistently express assigned traits across
elicitation tasks. We share their focus on agent consistency, but examine
instance-level decisions produced by a complete two-agent procedure. We move
the multiplicity question from model replacement to procedure replacement and
validate distinguishing cases with fresh runs.

\section{Experimental Setup}

\paragraph{Controlled testbed.}
The testbed contains 200 job postings, each paired with one constructed
r\'esum\'e in three intended-fit strata, for 600 application pairs. Strong
r\'esum\'es cover most required skills at the requested depth; borderline
r\'esum\'es deliberately combine matched and missing requirements; weak
r\'esum\'es come from mismatched functional domains. A fixed, LLM-free scorer
checked the intended strata, with 600/600 agreement after regeneration.
Regeneration was triggered only by scorer disagreement, never by a one-call or
two-agent outcome. The strata support controlled comparison and are not
qualification ground truth (Appendix~\ref{app:testbed}).

\paragraph{Procedures.}
Both procedures assess required skills, experience, and basic qualifications,
and both represent employer- and candidate-side perspectives in their decision
scores, but they combine those perspectives differently. The
\textbf{one-call procedure} gives one model joint access to both documents. In
a single call, it identifies job requirements, finds supporting r\'esum\'e
evidence, emits an employer-perspective score and a candidate-perspective
score, and returns \pass{} when their stored mean is at least .75. Its final
readout is therefore compensatory: a higher score on one perspective can offset
a lower score on the other.

The \textbf{two-agent procedure} assigns the roles and source documents
separately. The candidate-side agent sees the r\'esum\'e and responds only with
evidence grounded in it; the employer-side agent sees the posting, evaluates
fit, and probes unresolved requirements. After a fixed employer opening, turns
alternate candidate-then-employer. Both agents see the message history but not
the other side's source document, so job criteria and r\'esum\'e evidence must
cross through the exchange. An engine-level rule requires pending questions to
be answered before a new topic is introduced.

The evolving perspective scores feed a deterministic, non-compensatory rule
layer rather than a mean: it uses a .75 principal commitment condition with
.65 sustained-floor variants, and terminates on commitment, rejection,
divergence, stall, or the round limit. \pass{} is unavailable before round 3.
Eight rounds form the soft cap, extendable to at most 12. On the repeated borderline pool, one-call uses exactly one model
call per decision; two-agent sessions use a median of eight and a mean of 9.31
calls. Prompts and decision rules were fixed before runs A--C.

Before any model call, the two-agent procedure also applies a deterministic
\emph{pre-exchange skill-overlap gate}: if fewer than 10\% of normalized
required-skill names occur in the r\'esum\'e's declared skill list, the pair is
recorded as \fail{} without an exchange. Across the testbed, this gate stops
203/600 pairs: 0 strong, 9 borderline, and 194 weak. One-call screening has no
corresponding pre-exchange gate and scores every pair it is run on. For paired
borderline analyses, we
remove the same nine gated IDs from one-call results to define a common
191-pair pool.

\paragraph{Runs and measures.}
Runs A--C execute each procedure three times on the 191 borderline pairs under
\gpt{} and \opus{}; Run A additionally covers all strong and weak pairs. A pair
is \emph{persistent} if it passes all three runs and \emph{mixed} if it passes
one or two. Three-run borderline pass rates count pass instances over the
common pool (573 decisions per procedure and model); Table~\ref{tab:complete}
instead reports the single Run-A readout on all 200 borderline pairs. Because
the native one-call rule passes far fewer borderline applications, we
descriptively re-read its stored scores at the cutoffs closest to the two-agent
pass volumes (.590 \gpt{}; .575 \opus{}). The model is not rerun, and two-agent
outcomes are not re-thresholded.

Fresh runs use discovery-set membership at these readouts. Runs D--F execute
both procedures on 36 shared and 20 two-agent-only application--model cases.
Runs G--I separately execute only one-call screening on the same 36 shared and
35 one-call-only cases. ``Only'' means persistent under one procedure but below
3/3 under the other, not necessarily 3/3 versus 0/3. The endpoint is whether
the originating procedure again returns 3/3 \pass{}. Cohort labels are never
shown to the models.

\section{Findings}

\subsection{Two-agent screening changes decisions in both directions}
\label{sec:f1}

Run A is the only discovery run covering all 600 pairs. As
Table~\ref{tab:complete} shows, two-agent screening removes some one-call
passes and adds some one-call failures, so net differences understate changed
membership.

\begin{table}[t]
\centering
\footnotesize
\setlength{\tabcolsep}{2.5pt}
\begin{tabular}{llrrrrr}
\toprule
& & \multicolumn{2}{c}{\pass{}} & \multicolumn{3}{c}{Run-A disagreement} \\
\cmidrule(lr){3-4}\cmidrule(lr){5-7}
Model & Stratum & One & Two & Two rejects & Two adds & Net \\
\midrule
\gpt{}  & Strong     & 192 & 186 &  6 &  0 & $-6$ \\
\gpt{}  & Borderline &   8 &  50 &  0 & 42 & $+42$ \\
\gpt{}  & Weak       &   0 &   0 &  0 &  0 & 0 \\
\addlinespace[1pt]
\opus{} & Strong     & 189 & 184 & 10 &  5 & $-5$ \\
\opus{} & Borderline &  15 &  29 &  5 & 19 & $+14$ \\
\opus{} & Weak       &   0 &   0 &  0 &  0 & 0 \\
\bottomrule
\end{tabular}
\caption{Run-A outcomes for each 200-pair stratum. ``Two rejects'' counts
applications with one-call \pass{} and two-agent \fail{}; ``Two adds'' counts
the reverse. These are single-run disagreements, not the persistent-set
membership used in Sections~\ref{sec:f2} and~\ref{sec:f3}. Weak rows are descriptive because the two-agent
pre-exchange skill-overlap gate stops 194/200 pairs before exchange.}
\label{tab:complete}
\end{table}

Under \gpt{}, two-agent screening removes six strong applications and adds 42
borderline applications. Under \opus{}, disagreement runs both ways within both
strata: 10 strong applications leave and five enter, while five borderline
applications leave and 19 enter. The \gpt{} strong decrease is nominally
significant ($p=.031$), the \opus{} decrease is not ($p=.302$), and both
borderline increases are supported by exact paired tests
($p<5\times10^{-13}$; $p=.0066$). A uniformly lower one-call cutoff can add
applications but cannot remove existing passes, so these bidirectional changes
rule out simple uniform relaxation. Because the procedures differ in role
separation, \mbox{aggregation}, interaction, compute, and gating, this result is an
end-to-end procedural contrast rather than an attribution to dialogue alone.

We re-executed all 21 Run-A strong disagreements at least twice under both procedures.
Six remain oppositely classified across every execution, including cases in
both directions under \opus{}. This selected follow-up measures durability rather
than prevalence, but confirms recurring counterexamples to a uniformly relaxed
one-call rule (Appendix~\ref{app:complete}).

\subsection{Similar pass volumes do not mean the same applications}
\label{sec:f2}

The native procedures operate at very different pass volumes, which also
inflates their apparent difference in mixed-case rate. Re-reading the stored
one-call scores near the two-agent volumes reduces the mixed-case gap from
13.6 to 6.8 percentage points for \gpt{} and from 8.4 to 3.7 for \opus{}
(Table~\ref{tab:aligned}). The native mixed-case contrast should therefore
not be interpreted as an architectural property.

\begin{table}[H]
\centering
\footnotesize
\setlength{\tabcolsep}{2.7pt}
\begin{tabular*}{\columnwidth}{
  @{\extracolsep{\fill}}llrrrrr@{}
}
\toprule
Model & Readout &
\multicolumn{2}{c}{\shortstack{Pass\\instances}} &
\multicolumn{2}{c}{\shortstack{Mixed\\cases}} &
$p$ \\
\cmidrule(lr){3-4}\cmidrule(lr){5-6}
& & One & Two & One & Two & \\
\midrule
\gpt{ }  & Native  &  26 & 150 &  2 & 28 & $<.001$ \\
\gpt{ }  & Matched & 150 & 150 & 15 & 28 & .053 \\
\addlinespace[1pt]
\opus{} & Native  &  37 &  92 &  5 & 21 & .002 \\
\opus{} & Matched & 102 &  92 & 14 & 21 & .26 \\
\bottomrule
\end{tabular*}
\caption{Native and similar-volume readouts on the common 191-pair pool.
Pass instances are counted over 573 decisions and mixed cases over 191
three-run histories. Matched rows change only the stored one-call cutoff
and are descriptive.}
\label{tab:aligned}
\end{table}

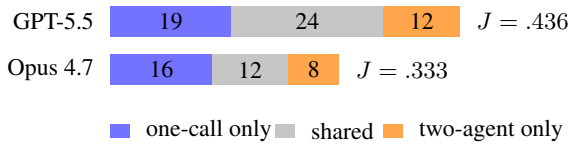
\begin{figure}[H]
\centering
\begin{tikzpicture}[x=0.084cm,y=0.58cm,font=\footnotesize]
  \node[anchor=east] at (-1,2.2) {\gpt{}};
  \fill[blue!55] (0,1.85) rectangle (19,2.55);
  \fill[gray!45] (19,1.85) rectangle (43,2.55);
  \fill[orange!70] (43,1.85) rectangle (55,2.55);
  \node at (9.5,2.2) {19}; \node at (31,2.2) {24}; \node at (49,2.2) {12};
  \node[anchor=west] at (56,2.2) {$J=.436$};
  \node[anchor=east] at (-1,1.1) {\opus{}};
  \fill[blue!55] (0,0.75) rectangle (16,1.45);
  \fill[gray!45] (16,0.75) rectangle (28,1.45);
  \fill[orange!70] (28,0.75) rectangle (36,1.45);
  \node at (8,1.1) {16}; \node at (22,1.1) {12}; \node at (32,1.1) {8};
  \node[anchor=west] at (37,1.1) {$J=.333$};
  \fill[blue!55] (0,-0.45) rectangle (3,-0.15);
  \node[anchor=west] at (4,-0.30) {one-call only};
  \fill[gray!45] (26,-0.45) rectangle (29,-0.15);
  \node[anchor=west] at (30,-0.30) {shared};
  \fill[orange!70] (43,-0.45) rectangle (46,-0.15);
  \node[anchor=west] at (47,-0.30) {two-agent only};
\end{tikzpicture}
\caption{Membership among applications that pass all three discovery runs at
the similar-volume one-call cutoffs.}
\label{fig:persistent}
\end{figure}

Rate matching still does not make the procedures interchangeable. At the
similar-volume readouts, run-level \pass{}-set Jaccard is .741--.865 across
\gpt{} one-call reruns and .656--.707 across \gpt{} two-agent reruns, but only
.347--.429 across procedures; the \opus{} ranges are .725--.811, .568--.750, and
.348--.408. Every within-procedure rerun pair therefore overlaps more than
every cross-procedure pair. For persistent selections, \gpt{} has 43 one-call and
36 two-agent passes, sharing 24; \opus{} has 28 and 20, sharing 12
(Figure~\ref{fig:persistent}). The resulting Jaccards are .436 and .333,
leaving 31 \gpt{} and 24 \opus{} persistent selections specific to one procedure.

Scanning every distinct stored one-call mean-score cutoff still cannot
reconstruct the two-agent persistent set: the best Jaccard is .436 for \gpt{} and
.361 for \opus{} (Appendix~\ref{app:sweep}). Alternative readouts of the two
component scores were not swept. Thus two-agent screening is not another operating point on the stored
one-call mean ranking.

\subsection{Recurrence differs within discovery-selected cohorts}
\label{sec:f3}

Overall fresh recurrence is similar: 45/56 (80.4\%) of the complete two-agent
persistent set and 60/71 (84.5\%) of the complete one-call set again pass all
three fresh runs. The results therefore do not support a general claim that
two-agent screening is less repeatable. The difference lies in how each total
is composed (Table~\ref{tab:fresh}).

\begin{table}[H]
\centering
\footnotesize
\begin{tabular*}{\columnwidth}{
  @{\extracolsep{\fill}}lccc@{}
}
\toprule
Origin / model & Shared & Origin only & Gap \\
\midrule
Two-agent (D--F) & 34/36 (94\%)  & 11/20 (55\%) & $-39$ pp \\
\quad \gpt{ }        & 24/24 (100\%) &  7/12 (58\%) & $-42$ pp \\
\quad \opus{}       & 10/12 (83\%)  &  4/8 (50\%)  & $-33$ pp \\
\addlinespace[1pt]
One-call (G--I)   & 29/36 (81\%)  & 31/35 (89\%) & $+8$ pp \\
\quad \gpt{ }         & 19/24 (79\%)  & 17/19 (89\%) & $+10$ pp \\
\quad \opus{}        & 10/12 (83\%)  & 14/16 (88\%) & $+4$ pp \\
\bottomrule
\end{tabular*}
\caption{Fresh 3/3 recurrence within discovery-selected cohorts,
by originating procedure and model. ``Origin only'' denotes persistence under
the originating procedure but below 3/3 under the other.}
\label{tab:fresh}
\end{table}

In D--F, two-agent recurrence falls from 34/36 shared to 11/20 two-agent-only
cases. The contrast is clear for \gpt{} (24/24 versus 7/12; Fisher $p=.0021$) but
uncertain for \opus{} (10/12 versus 4/8; $p=.161$). In the separate G--I block,
one-call recurrence is 29/36 for shared and 31/35 for one-call-only cases, so
the corresponding cohort does not show the same descriptive decline. This
comparison argues against a simple symmetric summary but does not establish an
architectural asymmetry: the selected cohorts may differ in latent difficulty,
and their discovery selection conditions are not equally diagnostic. We
therefore interpret the contrast as a property of these discovery cohorts, not
as evidence that two-agent screening is generally less reliable.

\section{Implications for Agent Evaluation}
When r\'esum\'e screening moves from a single call to an exchange between
candidate-side and employer-side agents, the object of evaluation becomes the
complete decision procedure, not merely the quality of the dialogue it
produces. Our results motivate three levels of measurement, none of which is
sufficient on its own.

\paragraph{Volume.}
How many applications advance? Because procedures may operate at different
effective thresholds, pass-rate differences alone cannot show whether a new
procedure reallocates access rather than merely operating at a different
threshold; rate alignment is needed to distinguish these descriptions.

\paragraph{Membership.}
Which applications advance at similar pass volumes? The two procedures selected
substantially different sets even when their pass volumes were similar, and no
threshold on the one-call score recovered the set consistently selected by
two-agent screening. The evaluated two-agent procedure therefore cannot be treated simply as a new operating point on the one-call ranking.

\paragraph{Recurrence.}
Do the selections that distinguish the procedures appear again when the
pipeline is re-executed? Within the discovery-selected cohorts evaluated in
fresh runs, two-agent-only selections recurred less often than selections
shared by both procedures. This selected-cohort result shows why direct
re-execution is informative, but it does not establish that two-agent
screening is generally less reliable.

\noindent
Reporting all three levels matters most when selection controls access. A
procedure may advance the expected number of applications while changing who
receives review, and some of those procedure-specific changes may not recur;
aggregate pass rates reveal neither.

\section{Conclusion}

We study what happens when the first gate in hiring moves from a static,
one-call judgment to an exchange between candidate-side and employer-side
agents. In our testbed, two-agent screening reallocates access rather than
merely changing pass volume. Similar totals do not recover the same
applications, and some procedure-specific selections do not recur in fresh
executions. These results do not establish which procedure makes better hiring
decisions; they show that aggregate agreement is insufficient evidence of
procedural equivalence. Evaluating agent-mediated screening therefore requires
application-level comparison and direct rerun validation, not pass rates alone.

\section*{Limitations}

\paragraph{Agent design.}
The candidate-side agent sees only the r\'esum\'e, so it can surface evidence
already present but cannot draw on work samples, portfolios, preferences, or
explanations that a real applicant might provide. The employer-side agent sees
only the public job posting, without access to team needs, internal evaluation
rubrics, or compensation constraints. Role prompts, communication rules,
stopping conditions, and decision thresholds are also fixed. Richer
information on either side and alternative interaction designs remain
unexplored; our findings characterize the evaluated procedures rather than
agent-mediated screening in general.

\paragraph{Procedure-level comparison.}
One-call and two-agent screening differ simultaneously in information access,
role separation, interaction structure, inference budget, and decision rules.
The one-call readout averages two perspective-specific scores, whereas the
two-agent procedure uses a non-compensatory rule layer and a two-agent-only
pre-exchange skill-overlap gate. We evaluate these as complete procedures and
do not isolate the contribution of dialogue, role separation, additional
inference, or any individual rule. A compute-matched iterative single-agent
baseline---matched on calls or tokens and evaluated under a pre-specified
decision rule---together with component-level ablations is needed for such
attribution. Because pairs stopped by the pre-exchange skill-overlap gate
terminate before any model call, the existing logs contain no trajectories
from which a gate-off counterfactual can be reconstructed; evaluating that
ablation requires new executions.

\paragraph{Rate matching.}
Rate matching changes only the one-call readout: we reapply alternative cutoffs
to stored one-call scores, choosing cutoffs to match the observed two-agent
pass volumes. The two-agent procedure remains at its native configuration,
because changing its commitment and stopping rules could alter the preceding
dialogue and would require new executions. The one-call procedure can therefore be examined
across operating points, whereas the two-agent procedure is evaluated at only
one operating point. The comparison is descriptive rather than symmetrically
tuned.

\paragraph{Fresh validation.}
Fresh runs cover discovery-selected persistent cases rather than the full
borderline set. Runs D--F and G--I evaluate different targeted cohorts, so we
do not treat their contrast as a formal between-procedure interaction test.
Because cases that were not persistent during discovery were not re-executed,
we cannot estimate full-pool discovery-to-fresh overlap.

\paragraph{Testbed and model scope.}
The r\'esum\'es are constructed from one job-posting collection, and the
construction strata are design categories rather than validated labor-market
labels. Because the testbed contains no real candidates or downstream
interview, offer, or job-performance outcomes, we cannot determine whether
either procedure is more accurate, fair, or beneficial. \gpt{} generated the
r\'esum\'es and also serves as one evaluator, creating potential
generator--evaluator dependence; evaluation with \opus{} does not fully remove
this concern. Provider-side sampling could not be fully controlled, and the
magnitude and direction of some results differ across models. Results may
therefore not generalize beyond the evaluated job collection, models, prompts,
thresholds, and executions.

\section*{Ethical Considerations}
Hiring is high stakes. This work evaluates screening procedures using
constructed r\'esum\'es and public job postings; it involves no real
applicants or hiring decisions. It does not recommend deployment, establish
that any application is qualified, or claim that exchanges between agents can
substitute for communication with actual candidates. Responsible deployment
would require job-related validation, human oversight, a meaningful appeal
process, privacy and accessibility protections, and demographic and disability
evaluation. Depending on the deployment context, relevant regulatory
frameworks may include New York City Local Law 144 and the EU AI Act
\citep{nyc2021ll144,europeanunion2024aiact}.

The constructed r\'esum\'es use names from a fixed placeholder list and
contain no explicit demographic attributes. We perform no demographic
analysis. Indirect signals such as institution prestige, geography, and
employment gaps were not controlled, so the testbed is unsuitable for
fairness or disparate-impact claims.

\section*{Acknowledgments}

We thank the anonymous reviewers for their comments and suggestions.
The generative AI tool (ChatGPT) was used for language polishing and improving the readability of the paper.

\bibliography{custom}

\appendix

\section{Testbed Construction and Protocol}
\label{app:testbed}

The testbed uses 200 unique posting identifiers from the \textit{LinkedIn Job
Postings (2023--2024)} dataset \citep{koneru2024linkedin}, downloaded on March
14, 2026 under CC BY-SA 4.0. Each posting is paired with one constructed
r\'esum\'e in each intended-fit stratum. Strong r\'esum\'es cover at least
80\% of required skills at the stated minimum depth or one level above, with
experience and salary inside the posting's range. Borderline r\'esum\'es cover
about 60\% at the minimum depth or one level below; omitted requirements are
excluded from all fields, and postings rotate a skill, seniority, or salary
mismatch. Weak r\'esum\'es have under 30\% required/preferred skill-name
overlap and a clear domain or seniority mismatch.

A fixed, LLM-free arithmetic scorer combines skill coverage, seniority, salary,
and location fit and assigns strong, borderline, or weak construction labels.
A r\'esum\'e was regenerated only when this scorer disagreed with the intended
stratum; weights and cutoffs were unchanged, and no regeneration depended on a
one-call or two-agent outcome. The final testbed has 200 pairs per stratum and
600/600 agreement between intended and scored labels. These labels describe
designed posting fit rather than candidate quality.

The one-call procedure makes exactly one model call with joint access to both
documents and applies the stored mean-score rule. The two-agent procedure
begins with a posting-specific fixed employer message, then alternates
candidate and employer model turns. The employer-side agent uses the posting to
state criteria and probe unresolved requirements; the candidate-side agent
responds using only r\'esum\'e-grounded evidence. Each agent sees the full
message history and its own source information, but not the other side's
document. Engine-level instructions require direct answers to pending
questions. The state begins at .40. The fixed rule layer uses a .75 principal
commitment condition, sustained-floor variants at .65, and independent
failure conditions. Commitment is suppressed before round 3, but rejection may
terminate earlier. Eight rounds form the soft cap; sessions may extend to at
most 12 rounds, and the four-round stall condition is evaluated only after the
soft cap.

A two-agent-only \emph{pre-exchange skill-overlap gate} records \fail{}
without a model call when fewer than 10\% of normalized required-skill names
occur in the r\'esum\'e's declared skill list. It gates 0 strong, 9 borderline,
and 194 weak pairs (203/600 total), with identical IDs across models and runs.
One-call screening has no such pre-exchange gate; the nine borderline IDs are
excluded from its results only to define the common 191-pair analysis pool.
Records stopped by this gate contain the engine's .40/.40 initialization
values rather than measured model scores and are not treated as scored
failures. A separate \emph{evidence-confirmation rule} converts an employer
\pass{} into another probe until 80\% of required skills are evidence-confirmed;
it never directly produces \fail{} and is distinct from the pre-exchange gate.
Prompts, thresholds, routes, and stopping rules were fixed before runs A–C.

\section{Strong-Stratum Re-execution}
\label{app:complete}

The target manifest and persistence endpoint were frozen before collection.
Every Run-A strong-stratum disagreement was re-executed under both procedures
with no selective reruns: six application--model cases under \gpt{ } and fifteen
under \opus{}, the latter comprising ten one-call advances that two-agent rejected
and five one-call rejections that two-agent advanced. Each case received three
executions per procedure including Run A, except the \gpt{ } one-call arm, which
received four; persistence outcomes are identical under either count.

A disagreement persists when one-call returns the same decision in every
execution and two-agent the opposite in every execution. Three of six persist
under \gpt{}; under \opus{}, two of ten removals and one of five advances persist, so
both directions retain recurring cases. Of the fifteen that dissolve, thirteen
do so because the two-agent decision changes. The other two involve one-call
scores adjacent to .75. Because the cohort is selected on disagreement in Run
A, these proportions describe the durability of observed disagreements, not
the prevalence of disagreement in an independently repeated strong stratum.

\section{Cutoff Sweep and Run Frequencies}
\label{app:sweep}

The one-call mean-score cutoff is scanned over all distinct stored decision
points after rounding scores to six decimals. \gpt{} matches the two-agent pass
volume exactly at .590. For \opus{}, .575 gives 102 pass instances and the next
distinct point gives 80; neither matches the two-agent total of 92 because 22
instances are tied at .575.

In a five-point neighborhood around the reported cutoffs, \gpt{} mixed-rate gaps
range from 4.7--6.8 percentage points and \opus{} gaps from 3.1--7.3. Across the
full sweep, the maximum persistent-set Jaccards with the two-agent set are .436
and .361. Thus the similar-volume point is not selected to disadvantage
reconstruction: it is the \gpt{} maximum, while the \opus{} matched Jaccard (.333)
lies below its maximum. At the original .75 cutoff the one-call mixed rates
are 2/191 (1.0\%) for \gpt{} and 5/191 (2.6\%) for \opus{}, which give the 13.6 and
8.4 point native gaps reported in Section~\ref{sec:f2}; across the full sweep
the highest one-call mixed rates are 36/191 (18.8\%) and 32/191 (16.8\%).
Table~\ref{tab:freq} gives the full three-run pass
frequency distributions at the reported cutoffs.

\begin{table}[h]
\centering
\footnotesize
\setlength{\tabcolsep}{3.0pt}
\begin{tabular}{lrrrrr}
\toprule
Procedure & 0/3 & 1/3 & 2/3 & 3/3 & \pass{} inst. \\
\midrule
\gpt{} one-call at .590 & 133 & 9 & 6 & 43 & 150 \\
\gpt{} two-agent & 127 & 14 & 14 & 36 & 150 \\
\opus{} one-call at .575 & 149 & 10 & 4 & 28 & 102 \\
\opus{} two-agent & 150 & 10 & 11 & 20 & 92 \\
\bottomrule
\end{tabular}
\caption{Pass frequencies in the common $N=191$ pool.}
\label{tab:freq}
\end{table}

\section{Fresh-Run Details}
\label{app:fresh}

The target manifests and endpoints were frozen before fresh collection. Runs
D--F cover 36 shared and 20 two-agent-only application--model cases under both
procedures, producing 336 procedure-level records with no errors or selective
reruns. Runs G--I separately cover the same 36 shared and 35 one-call-only
cases under one-call screening, producing 213 successful records. The main-text
fresh table uses the within-block shared comparator for each originating
procedure: 34/36 versus 11/20 for two-agent in D--F, and 29/36 versus 31/35 for
one-call in G--I. D--F also reran one-call on the shared cohort and obtained
31/36 at the aligned cutoff; this is a separate execution from the G--I 29/36
result and is not used as the G--I within-block comparator.
Table~\ref{tab:transitions} gives the persistent-set transition mapping for the
two D--F cohorts.

\begin{table}[h]
\centering
\footnotesize
\setlength{\tabcolsep}{1.5pt}
\begin{tabular}{@{}lrrrrr@{}}
\toprule
Discovery & Shared & TA only & OC only & Neither & Total \\
\midrule
Shared & 29 & 5 & 2 & 0 & 36 \\
TA only & 1 & 10 & 2 & 7 & 20 \\
\bottomrule
\end{tabular}
\caption{Pooled transitions from discovery A--C to fresh runs D--F. TA and OC
denote two-agent and one-call persistent membership.}
\label{tab:transitions}
\end{table}

Of the 20 two-agent-only cases, 10 remain two-agent-only and one becomes
shared, yielding the 11 two-agent 3/3 recurrences in Table~\ref{tab:fresh};
seven leave both persistent sets. Of the 36 shared cases, 29 remain shared and
none leaves both sets. In a stricter post-hoc subset requiring discovery
two-agent 3/3 and one-call 0/3, fresh two-agent persistence is 7/10 for \gpt{} and
3/5 for \opus{}.

For G--I, fresh one-call persistence is 19/24 for \gpt{} shared versus 17/19 for
\gpt{} one-call-only cases, and 10/12 versus 14/16 for \opus{}. Pooled, these are
29/36 and 31/35. Twenty-six pair IDs occur under both model configurations, so
pooled percentages are descriptive rather than independent observations.

\paragraph{Item-level variation.}
Among borderline applications that pass at least once at the similar-volume
readouts, 43.8\% are mixed under two-agent screening versus 25.9\% under
one-call for \gpt{}, and 51.2\% versus 33.3\% for \opus{}. Conditioning on one-call
below 3/3 therefore selects more strongly for mixed two-agent histories than
the reverse condition. Shared latent difficulty together with unequal
item-level mixed rates could produce both fresh-block contrasts; the data do
not isolate an architectural asymmetry.

\paragraph{Model identifiers.}
All runs requested \texttt{gpt-5.5} and \texttt{claude-opus-4-7}.
Returned-model logging was added after runs A--C, so those runs carry no
recorded identifier. In D--F the \texttt{gpt-5.5} alias resolved to
\texttt{gpt-5.5-2026-04-23}, the only published snapshot of that model, which
was released before runs A--C; G--I requested it directly, and separate probes
on 2026-05-27 and 2026-06-15 returned the same value. The Anthropic endpoint
returns the alias string rather than a dated identifier, so no dated \opus{}
version is available for any run.
\end{document}